\documentclass{article}
\usepackage{amsmath,graphicx}
\usepackage{iclr2016_conference,times}

\usepackage{color}
\usepackage{url}

\usepackage[caption=false]{subfig}

\usepackage{chngcntr}
\counterwithout{table}{subsection}

\begin{document}

\title{Mediated Experts for Deep Convolutional Networks}

\author{Sebastian Agethen, Winston H. Hsu \\National Taiwan University\\
Graduate Institute of Networking and Multimedia}

\maketitle
%Iteration: 1
%Abstract fits just barely, possibly shorten by one line?
\begin{abstract}
We present a new supervised architecture termed Mediated Mixture-of-Experts (MMoE) that allows us to improve classification accuracy of Deep Convolutional Networks (DCN). Our architecture achieves this with the help of expert networks: A network is trained on a disjoint subset of a given dataset and then run in parallel to other experts during deployment. A \emph{mediator} is employed if experts contradict each other.
This allows our framework to naturally support incremental learning, as adding new classes requires (re-)training of the new expert only.
We also propose two measures to control computational complexity: An early-stopping mechanism halts experts that have low confidence in their prediction. The system allows to trade-off accuracy and complexity without further retraining. We also suggest to share low-level convolutional layers between experts in an effort to avoid computation of a near-duplicate feature set. 
We evaluate our system on a popular dataset and report improved accuracy compared to a single model of same configuration. 
\end{abstract}

\section{Introduction}
\label{sec:introduction}
Deep learning methods in general, and \emph{Deep Convolutional Neural Networks} (DCN) in particular, have seen a surge in popularity among researchers over the past decade or so. While the application of early DCN was limited to simple tasks such as hand-written digit recognition~\cite{lit:lecun}, modern state-of-the-art methods can meet or exceed human-level performance on far more complex tasks, including generic image classification~\cite{lit:paramrectifier,lit:alexnet,lit:googlenet}.
Two factors that have contributed to this development are the availability of large-scale image corpora like \emph{ImageNet}~\cite{lit:imagenet}, and the wide-spread availability of GPU-based computing hardware which renders computationally expensive model training procedures feasible.

\begin{figure}[ht]
	\centering
  \def\svgwidth{1\columnwidth} 
	 \scalebox{.8}{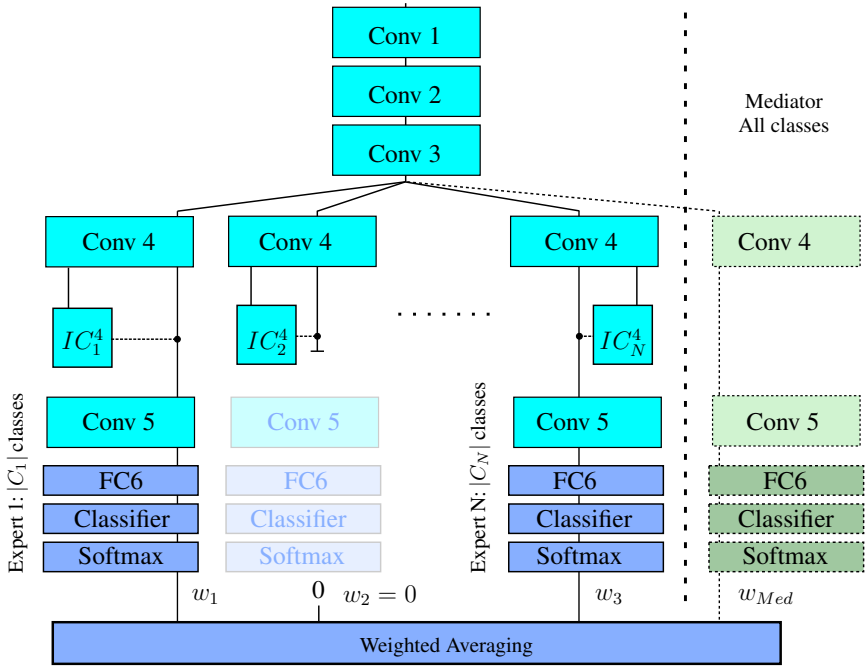}
	\caption{$N$ experts are trained on their corresponding \emph{superclasses} $C_i$. At inference time, \emph{confidence} is measured in \emph{confidence modules} $IC_i^j$. In this example, expert two is stopped at layer \texttt{Conv 4} due to low confidence and its activations are set to zero. A \emph{mediator} with full classifier improves accuracy of the system should more than one expert remain.}
	\label{fig:architecture}
\end{figure}

Experts systems are known to improve the classification accuracy of neural networks even further and have been studied extensively. 
Previous work on \emph{Mixture of Experts} (MoE) \cite{lit:hintonmixtureofexperts} is however flawed in two aspects:
First, computational complexity in MoE is a multiple of that of a traditional model. To this end, we propose early-stopping of experts. Our mechanism evaluates each expert's \emph{confidence} and is influenced by a hyper-parameter, allowing a trade-off between complexity and classification accuracy without the need for retraining.
Second, as recently pointed out in \cite{lit:incrementallearning}, training becomes more difficult as datasets are getting larger. Intuitively, we expect that an expert has extensive knowledge on a limited area of expertise, i.e., a subset of the training data, and thereby naturally avoids the issue.
Previous MoE systems are trained on full datasets, however, as difficulties arise when a decision is needed on which expert to trust. To remedy this issue, we introduce a \emph{mediator}, see Figure \ref{fig:architecture}. The mediator has a full classifier, allowing it to arbitrate conflicting predictions when multiple experts are confident.

In this paper we propose a new architecture which we name \emph{Mediated Mixture of Experts} (MMoE) that aims to increase the classification accuracy of a general DCN. We briefly discuss related research in Section~\ref{sec:related} before formally defining our method in Section~\ref{sec:branching}. We follow by summarizing the outcome of our experimental evaluation in Section~\ref{sec:evaluation} and give some concluding remarks in Section~\ref{sec:conclusion}.
\section{Related Work}
\label{sec:related}

% \textbf{TODO: Relate this work to previous work (Requirement Submission)}
%
A great deal of literature on deep learning and in particular on DCN has been published in recent years.
Image classification with deep learning has considerably profited from the availability of large datasets (in terms of both number of classes and images). In this paper, our choice falls on the popular ImageNet \cite{lit:imagenet} dataset. Both a 22K classes and a 1K classes version exist of this dataset, we evaluate our system on the latter.

\emph{AlexNet} by Krizhevsky \textit{et al}.~\cite{lit:alexnet} achieved a significant increase in classification performance over traditional methods on ImageNet, i.e., an top-1 accuracy of 62.5\%, showing the potential of deep learning for large datasets such as ImageNet. Their success is arguably founded on three factors: Depth, Rectified Linear Units (ReLU) as non-linearity and data augmentations. We use \emph{CaffeNet}, a variation of this model, as a baseline to evaluate our system.

A hierarchical architecture for incremental learning was presented in \cite{lit:incrementallearning}. The authors suggested to use a \textit{Branching} layer to determine the \textit{superclass} a specific problem belongs to. Following this, a leaf model provided fine-grained classification. We follow their work insofar that we also propose to use \textit{experts} (leaf models) for fine-grained classification. However, we avoid branching as it introduces a conditional error that cannot be recovered. Branching error also poses a scalability issue, as it grows with the number of superclasses. Finally, the prediction process in their work has large redundancy: Full models are used for both branching prediction and fine-grained prediction. %Many features of the former could be reused however.

Mixture of experts (MoE), as first proposed in \cite{lit:hintonmixtureofexperts}, have been well-known for a while. A large deal of work has been done in this area and an overview can be found in \cite{lit:twentyyearsofmoe}. Generally, a number $N$ of expert networks are trained together on a dataset. The experts learn discriminant features and thereby improve the overall accuracy of the system. While this can be used to improve classification results, the drawback lies in the parallel execution of all $N$ experts, which results in longer runtime. In contrast to our proposed method, experts are trained on the full dataset and have a complete classifier. This is disadvantageous for incremental learning, a scenario in which new classes are added over time, as all experts need to be retrained on the new data.

The work of \cite{lit:transferablefeatures} discusses \emph{generality} and \emph{specialization} of convolutional features, giving insight into how deep learning works. The same phenomenon has also been discussed in \cite{lit:visualizedcns} in the context of network visualization. In particular, both works show that the first few layers of a DCN produce \emph{general} Gabor-like features, i.e., lines and blobs. It is only in the higher layers that these features become more \emph{class-specific}.
Inspired by this fact, instead of $N$ times producing nearly identical general features, we can compute a single set of these features. 
\section{Proposed Method}
\label{sec:branching}
When tackling a problem, a divide-and-conquer approach can often be helpful: In deep learning, we can train experts on small problems such that the accuracy improves. This has been used in MoE \cite{lit:hintonmixtureofexperts}: Experts are trained competitively in the hope that they automatically learn discriminant features. A gating network combines the prediction results.

Each expert's area of expertise can also be designed with help of prior knowledge. We can utilize two methods to gain this knowledge: Spectral Clustering (as done in \cite{lit:incrementallearning}) and explicit hierarchies (such as the ImageNet dataset provides), see also Sec.~\ref{sec:hierdata}. In both cases the $i$-th expert is then trained exclusively on the subset of data in that superclass, which we denote as $C_i$.

\subsection{Simple Branching model} % This should be shorter probably, to allow more in-depth discussion later
We briefly discuss a \emph{branching network} as in \cite{lit:incrementallearning}, which predicts the superclass of an image. The branching network is a DCN such as \emph{AlexNet} \cite{lit:alexnet} with the number of outputs in the classifier reduced to $N$, the number of superclasses. In the \emph{branching decision}, the expert corresponding to the highest activation is selected to then obtain a fine-grained prediction. 

We trained such a system with two slim\footnote{By reducing the number of neurons of both layers \texttt{FC6} and \texttt{FC7} to 512, in the hope to keep parameter complexity low.} experts, as seen in Table \ref{tbl:twoexperts}. Even though we used two experts only, classification accuracy suffered from large branching errors: Should the wrong expert be chosen, the error cannot be recovered. Nevertheless, we also see that single experts show superior performance, confirming that we are on the right track. The model we discuss in the following is an attempt at reducing branching errors.
\begin{table}
	\caption{Branching networks introduce additional error: We trained such a network with two experts on superclasses $C_1,C_2$ as in Sec.~\ref{sec:hierdata}. Due to branching, the average accuracy on the full dataset $C_1\cup C_2$ remains below the traditional model. However, experts on their own have encouragingly higher accuracy on their area of expertise than a traditional model, see columns $C_1,C_2$.
	Note that the accuracy shown here was achieved with a slim configuration of \emph{CaffeNet}, i.e., the FC layers were reduced from 4096 to 512 neurons.}
	\centering
	\small
	\begin{tabular}{|l|l|l|l|}
		\hline
		Name & $C_1 \cup C_2$ & Class $C_1$ & Class $C_2$ \\ \hline \hline
		Branching & 93.2 \% & 93.35\% & 93.04\%  \\ \hline
		Expert 1 & -- & \textbf{46.262 \%} & -- \\ \hline
		Expert 2 & -- & -- & \textbf{59.202\%} \\ \hline
		Average & 49.151\% & -- & -- \\ \hline \hline
		Baseline \cite{lit:alexnet} & \textbf{49.35} \% & 41.52 \% & 55.96 \% \\ \hline
	\end{tabular}

	\label{tbl:twoexperts}
\end{table}

\begin{figure}[h]
	\centering
	\subfloat[]{\includegraphics[width=.33\columnwidth]{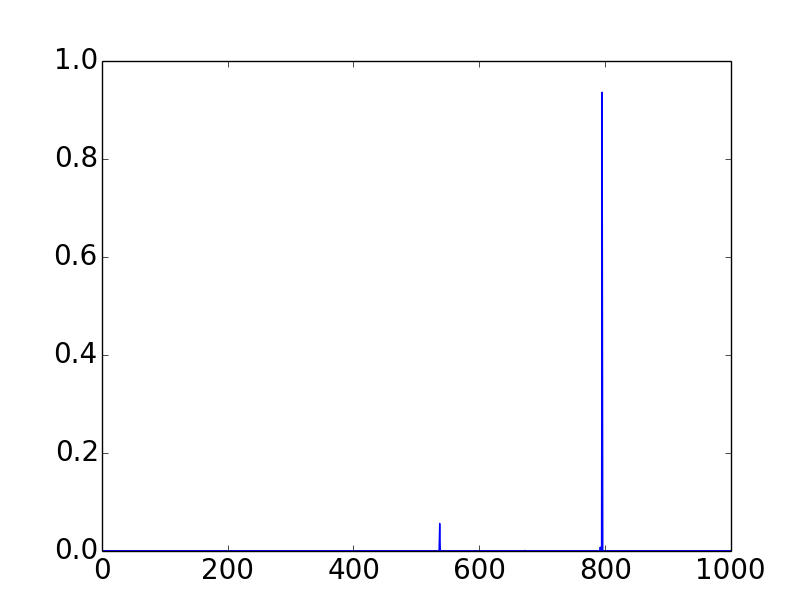}}
	\subfloat[]{\includegraphics[width=.33\columnwidth]{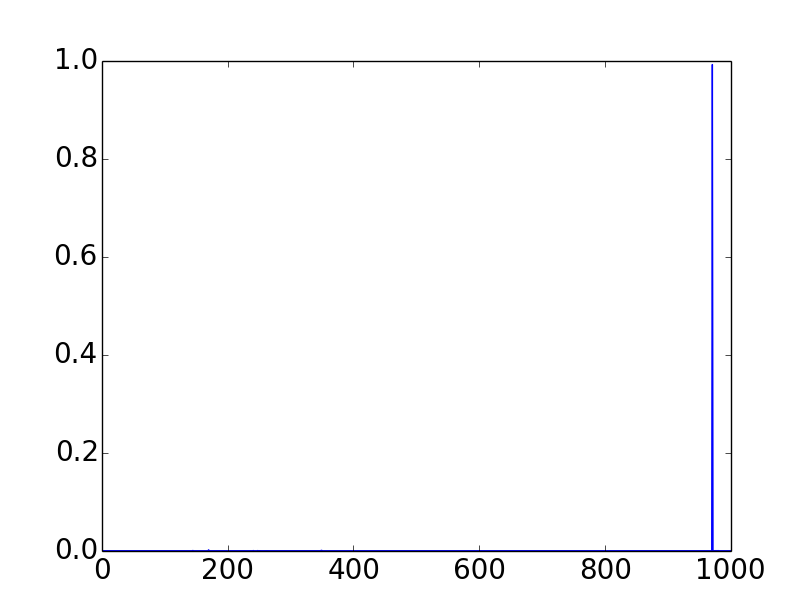}}
	\subfloat[]{\includegraphics[width=.33\columnwidth]{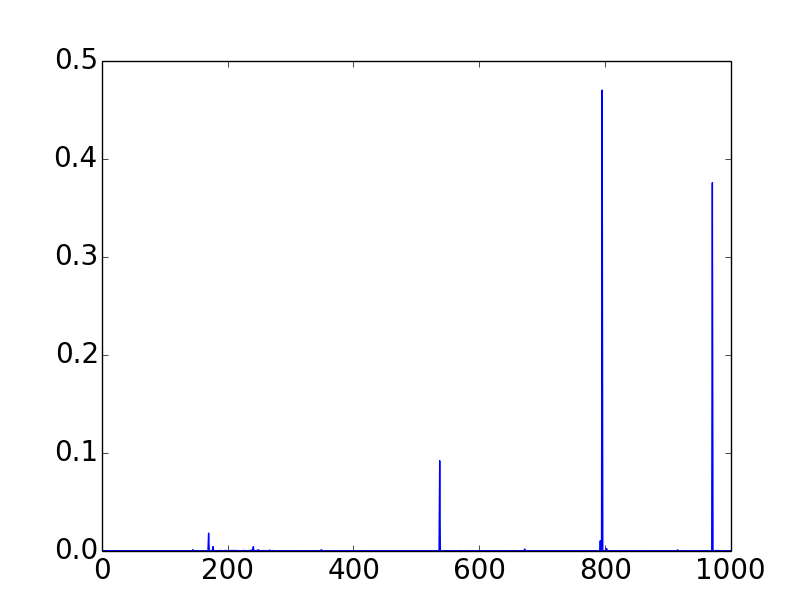}}
	\caption{Mediation process: Experts (a) and (b) both show strong opinions that contradict each other. Mediator (c) is able to solve the conflict (here in favor of expert (a)).}
	\label{fig:mediator}
\end{figure}

\subsection{Branched experts with early stopping}
We propose the following framework as depicted in Fig.~\ref{fig:architecture}: The input is passed through a number of convolutional layers to generate \textit{general} features, as defined in \cite{lit:transferablefeatures}. These layers are shared between all experts and could encompass the first two or three layers in the case of \emph{CaffeNet}. Higher layer features have a higher degree of specialization, and therefore need to be finetuned for each expert. Parallel execution of these layers is costly, however, in particular for large $N$. To this end we propose a \textit{confidence} module, which determines the confidence of whether the expert is able to solve the problem. In particular, an expert $i$ is deemed to have low confidence (in layer $j$) if for its given score $s_i^j$ and a threshold $T$:
\begin{equation}
	\max_{k\neq i} \left(s_k^j\right) - s_i^j \geq T, \quad	1 \leq i,k \leq N
	\label{eq:condition}
\end{equation}

In the following, we implement the confidence module simply as a fully connected layer with $N$ outputs, trained on the complete dataset with the original labels replaced by a superclass label $l_{C_i} \in\left[0,N-1\right]$. Note the placement of the confidence module: While lower layers have too general features to compute a reliable score, early stopping in higher layers diminishes the resource saving effect. We suggest an intermediate position such as after \texttt{Conv4}.

Let $\mathbf{u_{C_i}^{j}}$ be the activation of the confidence module in layer $j$ of the expert on $C_i$. The appropriate score is then simply the i-th component of the activation vector: $$s_i^j = \mathbf{u_{C_i,i}^j}$$
To support our decision for a simple threshold, we run a simple branching network (as described above) on a test set, and measure the magnitude of the left-hand side of Equation (\ref{eq:condition}). We then note whether the branching decision is correct or not. As is to be expected, bad branching decisions occur when this magnitude is small, while correct decisions often have higher confidence, see Table \ref{tbl:confidencebehavior}.

\begin{table}[t]
	\caption{Mean and standard deviation of confidence scores in cases of correct and incorrect branching decisions. Incorrect decisions very often imply a low confidence value, allowing us to decide whether additional experts need to be used to solve the task.}
	\centering
	\begin{tabular}{|l|l|l|}
	\hline
	Branching & Mean & Std.-dev. \\ \hline \hline
	Correct & 6.1 & 3.44 \\ \hline
	False & 1.8 & 1.67 \\ \hline
	\end{tabular}
	\label{tbl:confidencebehavior}
\end{table}

%\paragraph{Mediator}
\subsection{Mediator}
Given that the inequality in (\ref{eq:condition}) holds, an expert is deemed to have low confidence and stopped. His corresponding activations in the final layer are set to zero.
So far, the system described above performs slightly worse than a single model with same configuration. This is due to those instances, where more than one expert is active: An expert trained on a different superclass could ``mistake" the input for a particular fine-grained class of his own domain, leading to conflicting opinions. In these cases, we found it helpful to add a \emph{mediator}, that is, a slim model trained on all 1000 classes to arbitrate opinions, see Figure \ref{fig:mediator}. The resulting architecture then outperforms the single model baseline. 
Finally, the softmax probabilities of each expert are weighted and averaged: The slim 1000-classes mediator is only added when more than one expert is executed and weighted with $w_{Med}$. We set $w_{Med} = 0.6$ if executed or $w_{Med} = 0$ otherwise. Experts are weighted according to their confidence scores normalized to $[0,1]$ and subsequently scaled by $(1-w_{Med})$.

Our framework can also be used in the context of incremental learning. To add a new superclass $C_{N+1}$, we simply train a single expert on the new data. Following this, the confidence modules of other experts need to be fine-tuned to accommodate for the new superclass. We stress that this is significantly faster than training a full network, as the confidence module only encompasses a single layer with $N$ outputs, rendering it very shallow and slim. In our experiments, 3 epochs were found to be sufficient. In order to update the mediator network, we can simply add extra neurons to the output and finetune the network, see the Flat Increment technique in \cite{lit:incrementallearning}.

\section{Evaluation}
\label{sec:evaluation}
The evaluation is performed on the ImageNet 1K dataset. We begin by describing how the superclasses corresponding to each expert are formed.

\subsection{Superclass construction}
\label{sec:hierdata}
Superclasses can be defined in an automated or manual fashion: In the former case \emph{Spectral Clustering}, as was done in \cite{lit:incrementallearning}, is a suitable choice. Alternatively, we can traverse the hierarchy available with ImageNet and join several leaf classes that are conceptually related. This provides a convenient advantage in an incremental learning scenario: We simply train one or more new experts for new data. Contrary, when using Spectral Clustering, it may be necessary to retrain some of the existing experts as well.

%\subsection{Hierarchical dataset}
Due to space limitations, we limit ourselves to manually defined superclasses and $N=2$. We choose $C_1$ as ``artifact" (with synset: n00021939) and $C_2$ as all remaining classes, resulting in a split of 517 vs.~483 leaf classes with roughly 660K and 620K images each. In other words, this simulates adding ``artifact" data to an existing model. In the following, we refer to this as the \emph{Hierarchical set}.

\subsection{Results on Hierarchical set}
\textbf{Network configuration.} We evaluate our system on three configurations: First, a slim configuration of \emph{CaffeNet}, where both FC layers are reduced to 512 neurons. In our second configuration we drop layer \texttt{FC7} altogether, while reducing \texttt{FC6} to 512 neurons. Finally, our third configuration uses the default CaffeNet configuration with both fully connected layers containing 4096 neurons. All configurations were finetuned from the \emph{CaffeNet} model available with Caffe, see \cite{lit:caffealexnetmodel}.

\begin{figure}[h]
	\centering
	\includegraphics[width=.6\columnwidth]{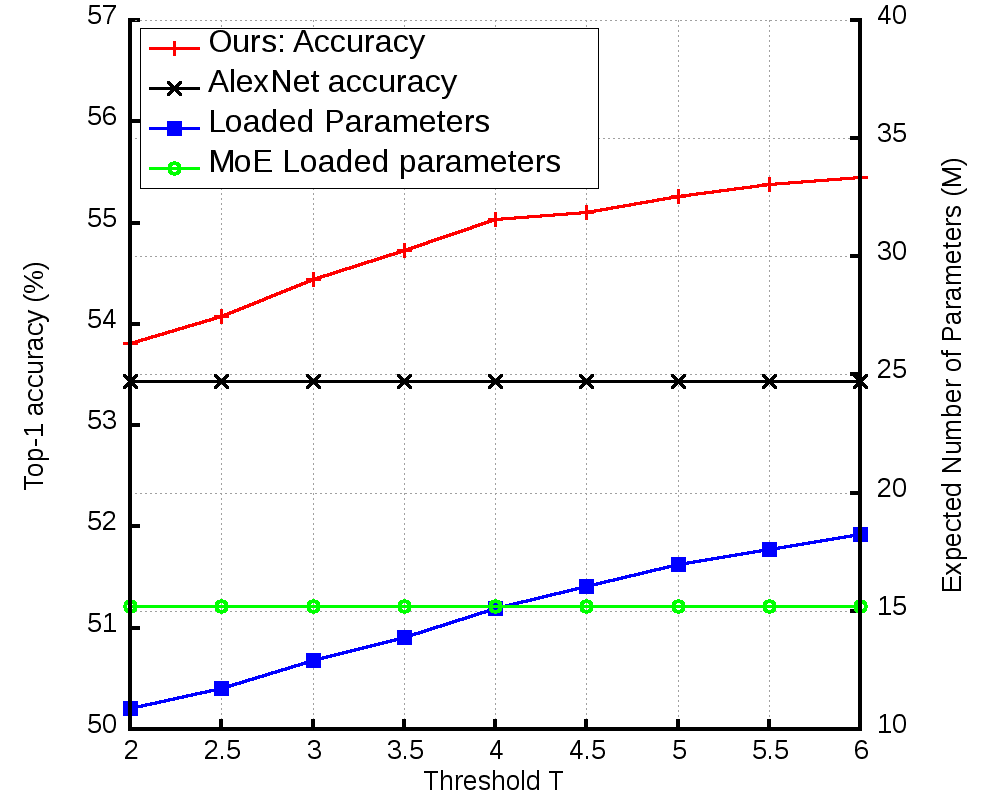}
	\caption{Top-1 accuracy of our proposed system in relation to threshold $T$, here on the slim seven layer configuration with the first three layers shared. The expected number of parameters to be loaded equals that of traditional MoE systems for $T=4$.}
	\label{fig:threshold}
\end{figure}

\textbf{System accuracy.} MMoE improves the top-1 classification accuracy in all three configurations. The result is strongly dependent on the value of threshold $T$, see Figure \ref{fig:threshold}. Classification accuracy peaks at $T=6$, for which our system outperforms the baseline over 2.8\% and 2.7\%\footnotemark[2] (in the slim and default configuration respectively). 

\begin{table}
	\caption{Top-1 accuracy, and stopping probabilities $p_1,p_2$ of both experts for $T=4$ under three different configurations.} 
	\small
	\centering
	\begin{tabular}{|l|l|l|l|}
		\hline
		Name & Acc. & $p_1$ & $p_2$  \\ \hline \hline
		MMoE (slim, 7 layers) & \textbf{0.5618} & 0.2213 & 0.1512 \\ \hline 
		MMoE (slim, 8 layers)& \textbf{0.5296} & 0.1805 & 0.1545 \\ \hline
		MMoE (default) & \textbf{0.5855} & 0.2225 & 0.17144  \\ \hline \hline
		MMoE (slim, 7 layers, shared) & 0.5361 & 0.2122 & 0.1801  \\ \hline
		Unmediated (slim, 7 layers, shared) & 0.5052 & 0.2122 & 0.1801 \\ \hline \hline
		Baseline (slim, 7 layers) & 0.5344 & -- & --  \\ \hline
		Baseline (slim, 8 layers) & 0.4948 & -- & --  \\ \hline
		Baseline \cite{lit:caffealexnetmodel} & 0.5584 & -- & --  \\ 
		 & (0.574) \footnotemark[2]& & \\ \hline
	\end{tabular}
	\label{tbl:results}
\end{table}
\footnotetext[2]{Using caffe's \texttt{test} command, we measured \emph{CaffeNet}'s top-1 accuracy at 55.84\% on the validation set, the reported accuracy is 57.4\% however.}

\begin{figure}[h]
	\begin{minipage}[t]{.5\textwidth}
			\includegraphics[width=\linewidth]{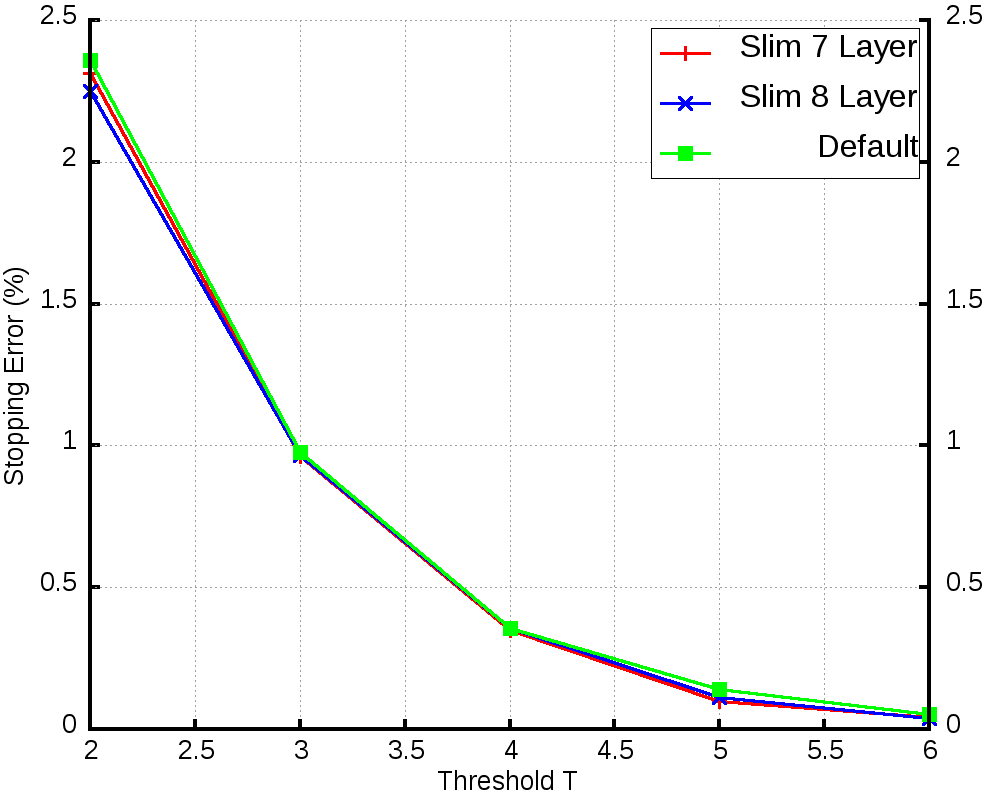}
			\caption{Probability that the true expert is falsely stopped in relation to threshold $T$. The error is significantly lower than the branching error in Section \ref{sec:branching} (6.8\%).}
			\label{fig:error}
	\end{minipage}\qquad
	\begin{minipage}[t]{.5\textwidth}
		\includegraphics[width=\linewidth]{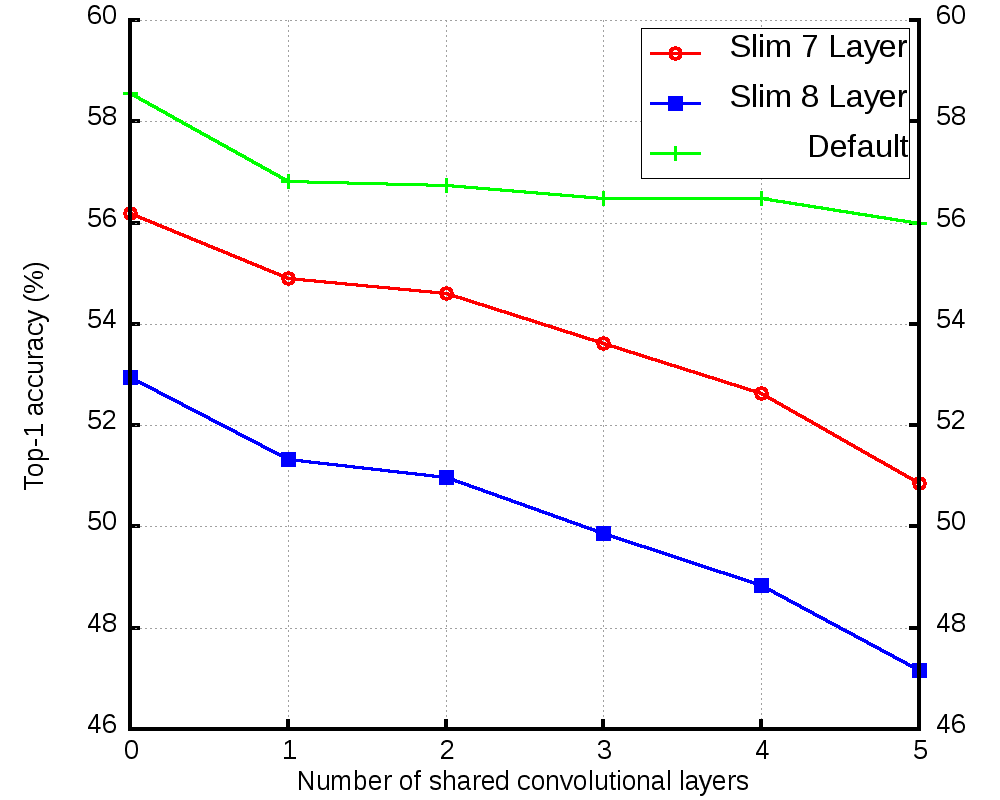}
		\caption{Performance of MMoE at $T=4$ with varying number of shared layers. The default configuration (CaffeNet) shows greater resilience as its performance only drops by 2 percent.}
		\label{fig:shared}
	\end{minipage}
\end{figure}

\textbf{Early stopping.} Experts are rarely stopped falsely, see Figure \ref{fig:error}. For $T\geq 3$, a true expert is incorrectly stopped in less than 1\% of all cases, an improvement over the branching error discussed in Section \ref{sec:branching}. Table \ref{tbl:results} shows the probability of how often an expert is stopped for all configurations.

\textbf{Mediator impact.} To show the importance of the Mediator, we train a slim configuration lacking mediation. As such, the system only reaches a top-1 accuracy of 50.52\%, about 3.1\% lower than the same configuration with mediator.

\textbf{Complexity vs.~Threshold.} The choice of $T$ also influences how often an expert is run. We show this by evaluating the average number of parameters that need to be loaded, see Figure \ref{fig:threshold}. For $T=4$, the expected number of parameters is close to the number of parameters when using traditional Mixture-of-Experts, that is, the complexity of $N$ experts alone. The number of computational operations, dominated by the Convolutional layers, can be significantly reduced by layer sharing as discussed earlier.

\textbf{Impact of sharing layers.} We also tested how sharing different numbers of convolutional layers affects performance. To achieve this, we finetuned the experts from the mediator, while keeping the lowest $k$ convolutional layers frozen, i.e., setting the learning rate to 0. We show our results for all three aforementioned configurations in Figure \ref{fig:shared}. Performance drops with larger number of shared layers, as is to be expected. The default configuration, having 4096 neurons in each fully connected layer, shows smaller impact: Accuracy drops by two percent going from no shared to fully shared layers, leading us to the assumption that the fully connected layers are able to mitigate the impact. The other two slim configurations however suffer a loss of more than 5\% when sharing all convolutional layers.

% \pagebreak
\section{Conclusion and Future Work}
\label{sec:conclusion}
We propose a mediated expert system for deep convolutional networks that enables experts to learn on small partitions of a training set, a case given in an incremental learning scenario. In detail, experts can be stopped early, where the stopping point is controlled by a single hyper-parameter, allowing to adapt to different circumstances in terms of availability of resources. Furthermore, we avoid the branching error that occurs in prior work when training on partitioned datasets.

In order to better underline the benefits of our proposal, two points are left for our future work. First, a more thorough evaluation is necessary that was not possible in the scope of this paper. Second, we believe that the mediator concept can be developed further in order to reduce computational complexity to a higher degree.

\bibliography{paper-iclr}
\bibliographystyle{iclr2016_conference}

\end{document}